%% file: paper.tex
\documentclass[10pt,twocolumn,letterpaper]{article}

\usepackage{cvpr}
\usepackage{times}
\usepackage{epsfig}
\usepackage{graphicx}
\usepackage{amsmath}
\usepackage{amssymb}

\usepackage{url}            
\usepackage{booktabs}       
\usepackage{amsfonts}       
\usepackage{nicefrac}       
\usepackage{microtype}      
\usepackage{enumitem}
\usepackage[dvipsnames]{xcolor}

\input{vcl-shortcuts.tex}
\newcommand{\ve}[1]{\mathbf{#1}} 
\newcolumntype{K}[1]{>{\centering\arraybackslash}p{#1}}
\newcommand{\cmark}{\Large \color{ForestGreen}\ding{51}}
\newcommand{\xmark}{\Large \color{red}\ding{55}}

\newcolumntype{x}[1]{>{\centering\arraybackslash}p{#1pt}}

\makeatletter\renewcommand\paragraph{\@startsection{paragraph}{4}{\z@}
	{.5em \@plus1ex \@minus.2ex}{-.5em}{\normalfont\normalsize\bfseries}}\makeatother

\usepackage[pagebackref=true,breaklinks=true,letterpaper=true,colorlinks,bookmarks=false]{hyperref}

\cvprfinalcopy

\def\cvprPaperID{***}

\ifcvprfinal\fi

\begin{document}

\title{Exploring Self-attention for Image Recognition}

\author{Hengshuang Zhao\\
CUHK\\
\and
Jiaya Jia\\
CUHK\\
\and
Vladlen Koltun\\
Intel Labs\\
}

\maketitle

\begin{abstract}
Recent work has shown that self-attention can serve as a basic building block for image recognition models. We explore variations of self-attention and assess their effectiveness for image recognition. We consider two forms of self-attention. One is pairwise self-attention, which generalizes standard dot-product attention and is fundamentally a set operator. The other is patchwise self-attention, which is strictly more powerful than convolution. Our pairwise self-attention networks match or outperform their convolutional counterparts, and the patchwise models substantially outperform the convolutional baselines. We also conduct experiments that probe the robustness of learned representations and conclude that self-attention networks may have significant benefits in terms of robustness and generalization.
\end{abstract}

\section{Introduction}
\label{sec:introduction}
\input{tex/introduction.tex}

\section{Related Work}
\label{sec:related}
\input{tex/related-work.tex}

\section{Self-attention Networks}
\label{sec:method}
\input{tex/method.tex}

\section{Experiments}
\label{sec:experiments}
\input{tex/experiments.tex}

\section{Conclusion}
\label{sec:conclusion}
\input{tex/conclusion.tex}

\balance

{\small
\bibliographystyle{ieee_fullname}
\bibliography{paper}
}

\end{document}

%% file: vcl-shortcuts.tex
\usepackage{times}
\usepackage{epsfig}
\usepackage{graphicx}
\usepackage{float}
\usepackage{wrapfig}
\usepackage{amsmath,amssymb,amsthm}
\usepackage{algorithm,algorithmicx,algpseudocode}
\usepackage{bm,xspace}
\usepackage{comment}
\usepackage{verbatim}
\usepackage{multirow}
\usepackage{balance}
\usepackage{url}
\usepackage{booktabs}
\usepackage{etoolbox,siunitx}
\usepackage{calc}
\usepackage{pifont,hologo}

\usepackage[super]{nth}
\usepackage{nicefrac}
\robustify\bfseries

\def\pp{\mathbf{p}}

\def\sss{\mathbf{s}}
\def\tt{\mathbf{t}}

\def\xx{\mathbf{x}}
\def\yy{\mathbf{y}}

\def\rR{\mathcal{R}}

\DeclareMathSymbol{@}{\mathord}{letters}{"3B}

\newcommand\timess{\mathbin{\!\times\!}}

\def\latex/{\LaTeX}
\def\bibtex/{\hologo{BibTeX}}


%% file: tex/introduction.tex
Convolutional networks have revolutionized computer vision. Thirty years ago, they were applied successfully to recognizing handwritten digits~\cite{LeCun1989}. Building directly on this work, convolutional networks were scaled up in 2012 to achieve breakthrough accuracy on the ImageNet dataset, outperforming all prior methods by a large margin and launching the deep learning era in computer vision~\cite{krizhevsky2012imagenet,Russakovsky2015}. Subsequent architectural improvements yielded successively larger and more accurate convolutional networks for image recognition, including GoogLeNet~\cite{szegedy2015going}, VGG~\cite{simonyan2014very}, ResNet~\cite{he2016deep}, DenseNet~\cite{huang2017densely}, and squeeze-and-excitation~\cite{hu2018squeeze}. These architectures in turn serve as templates for applications in computer vision and beyond.

All these networks, from LeNet~\cite{LeCun1989} onwards, are based fundamentally on the discrete convolution.
The discrete convolution operator $\ast$ can be defined as follows:
\begin{equation}
(F \ast k)(\pp) = \sum_{\sss + \tt = \pp} F(\sss) \, k(\tt).
\label{eq:regular}
\end{equation}
Here $F$ is a discrete function and $k$ is a discrete filter.
A key characteristic of the convolution is its translation invariance: the same filter $k$ is applied across the image $F$. While the convolution has undoubtedly been effective as the basic operator in modern image recognition, it is not without drawbacks. For example, the convolution lacks rotation invariance. The number of parameters that must be learned grows with the footprint of the kernel $k$. And the stationarity of the filter can be seen as a drawback: the aggregation of information from a neighborhood cannot adapt to its content.
Is it possible that networks based on the discrete convolution are a local optimum in the design space of image recognition models? Could other parts of the design space yield models with interesting new capabilities?

Recent work has shown that self-attention may constitute a viable alternative for building image recognition models~\cite{hu2019local,ramachandran2019stand}. The self-attention operator has been adopted from natural language processing, where it serves as the basis for powerful architectures that have displaced recurrent and convolutional models across a variety of tasks~\cite{vaswani2017attention,Devlin2018,dai2019transformer,Yang2019xlnet}. The development of effective self-attention architectures in computer vision holds the exciting prospect of discovering models with different and perhaps complementary properties to convolutional networks.

In this work, we explore variations of the self-attention operator and assess their effectiveness as the basic building block for image recognition models. We explore two types of self-attention. The first is \emph{pairwise} self-attention, which generalizes the standard dot-product attention used in natural language processing~\cite{vaswani2017attention}. Pairwise attention is compelling because, unlike the convolution, it is fundamentally a set operator, rather than a sequence operator. Unlike the convolution, it does not attach stationary weights to specific locations ($\sss$ in equation (\ref{eq:regular})) and is invariant to permutation and cardinality. One consequence is that the footprint of a self-attention operator can be increased (e.g., from a $3\timess 3$ to a $7\timess 7$ patch) or even made irregular without any impact on the number of parameters. We present a number of variants of pairwise attention that have greater expressive power than dot-product attention while retaining these invariance properties. In particular, our weight computation does not collapse the channel dimension and allows the feature aggregation to adapt to each channel.

Next, we explore a different class of operators, which we term \emph{patchwise} self-attention. These operators, like the convolution, have the ability to uniquely identify specific locations within their footprint. They do not have the permutation or cardinality invariance of pairwise attention, but are strictly more powerful than convolution.

Our experiments indicate that both forms of self-attention are effective for building image recognition models. We construct self-attention networks that can be directly compared to convolutional ResNet models~\cite{he2016deep}, and conduct experiments on the ImageNet dataset~\cite{Russakovsky2015}. Our pairwise self-attention networks match or outperform their convolutional counterparts, with similar or lower parameter and FLOP budgets. Controlled experiments also indicate that our vectorial operators outperform standard scalar attention. Furthermore, our patchwise models substantially outperform the convolutional baselines. For example, our mid-sized SAN15 with patchwise attention outperforms the much larger ResNet50, with a 78\% top-1 accuracy for SAN15 versus 76.9\% for ResNet50, with a 37\% lower parameter and FLOP count. Finally, we conduct experiments that probe the robustness of learned representations and conclude that self-attention networks may have significant benefits in terms of robustness and generalization.

%% file: tex/related-work.tex
\paragraph{Convolutional networks.}
Convolutional networks have come to dominate computer vision. More than two decades after their pioneering application to recognizing handwritten digits~\cite{LeCun1989}, ConvNets became mainstream after their successful application to image recognition on the ImageNet dataset~\cite{krizhevsky2012imagenet,Russakovsky2015}. A succession of increasingly powerful convolutional architectures for image recognition followed~\cite{szegedy2015going,simonyan2014very,he2016deep,huang2017densely,hu2018squeeze}. These serve as the basis for models developed for other computer vision tasks, such as semantic segmentation~\cite{long2015fully,chen2015semantic,yu2016multi,zhao2017pspnet} and object detection~\cite{girshick2014rcnn,girshick2015fastrcnn,ren2015faster,liu2016ssd}.

\paragraph{Self-attention.}
Self-attention models have revolutionized machine translation and natural language processing more broadly~\cite{vaswani2017attention,wu2019pay,Devlin2018,dai2019transformer,Yang2019xlnet}. This has inspired applications of self-attention and related ideas to image recognition~\cite{dai2017deformable,wang2017residual,hu2018squeeze,hu2018gather,zhao2018psanet,zhu2019deformable,hu2019local,bello2019attention,ramachandran2019stand}, image synthesis~\cite{zhang2018self,parmar2018image,brock2019large}, image captioning~\cite{xu2015show,yang2016stacked,chen2017sca}, and video prediction~\cite{jia2016dynamic,wang2017non}.

Until very recently, applications of self-attention in computer vision were complementary to convolution: forms of self-attention were primarily used to create layers that were used in addition to, to modulate the output of, or otherwise in combination with convolutions.
In channelwise attention models~\cite{wang2017residual,hu2018squeeze,hu2018gather}, attention weights reweight activations in different channels.
Other approaches~\cite{chen2017sca,woo2018cbam,fu2019dual} adopted both spatial and channel attention.
A number of methods learned to reweight convolutional activations or offset the taps of convolutional kernels~\cite{dai2017deformable,hu2018squeeze,wang2017residual,woo2018cbam,zhu2019deformable}, thus retaining the basic principle of convolutional feature construction. Others applied self-attention in specific modules that were appended to convolutional structures~\cite{wang2017non,zhao2018psanet}. Bello et al.~\cite{bello2019attention} combined convolutional and self-attention processing streams, but found that the global self-attention they used was not sufficiently powerful to replace convolutions entirely.
Jia et al.~\cite{jia2016dynamic} explored dynamic filter networks, which generalized convolutions, but the construction incurred significant memory and computational costs and was not scaled up to high-resolution images and larger datasets.

Most closely related to our work are the recent results of Hu et al.~\cite{hu2019local} and Ramachandran et al.~\cite{ramachandran2019stand}. One of their key innovations is restricting the scope of self-attention to a local patch (for example, $7\timess 7$ pixels), in contrast to earlier constructions that applied self-attention globally over a whole feature map~\cite{wang2017non,bello2019attention}. Such local attention is key to limiting the memory and computation consumed by the model, facilitating successful application of self-attention throughout the network, including early high-resolution layers. Our work builds on these results and explores a broader variety of self-attention formulations. In particular, our primary self-attention mechanisms compute a \emph{vector} attention that adapts to different channels, rather than a shared scalar weight. We also explore a family of patchwise attention operators that are structurally different from the forms used in~\cite{hu2019local,ramachandran2019stand} and constitute strict generalizations of convolution. We show that all the presented forms of self-attention can be implemented at scale, with favorable parameter and FLOP budgets.

%% file: tex/method.tex
In convolutional networks for image recognition, the layers of the network perform two functions. The first is feature aggregation, which the convolution operation performs by combining features from all locations tapped by the kernel. The second function is feature transformation, which is performed by successive linear mappings and nonlinear scalar functions: these successive mappings and nonlinear operations shatter the feature space and give rise to complex piecewise mappings.

One observation that underlies our construction is that these two functions -- feature aggregation and feature transformation -- can be decoupled. If we have a mechanism that performs feature aggregation, then feature transformation can be performed by perceptron layers that process each feature vector (for each pixel) separately. A perceptron layer consists of a linear mapping and a nonlinear scalar function: this pointwise operation performs feature transformation. Our construction therefore focuses on feature aggregation.

The convolution operator performs feature aggregation by a fixed kernel that applies pretrained weights to linearly combine feature values from a set of nearby locations. The weights are fixed and do not adapt to the content of the features. And since each location must be processed with a dedicated weight vector, the number of parameters scales linearly with the number of aggregated features. We present a number of alternative aggregation schemes and construct high-performing image recognition architectures that interleave feature aggregation (via self-attention) and feature transformation (via elementwise perceptrons).

\subsection{Pairwise Self-attention}
\label{sec:pairwise}

We explore two types of self-attention. The first, which we refer to as \emph{pairwise}, has the following form:
\begin{equation}
	\label{eq:pairwise}
	\ve{y}_{i} = \sum_{j \in \rR(i)} \alpha(\ve{x}_i, \ve{x}_j) \odot \beta(\ve{x}_{j}),
\end{equation}
where $\odot$ is the Hadamard product, $i$ is the spatial index of feature vector $\xx_i$ (i.e., its location in the feature map), and $\rR(i)$ is the local footprint of the aggregation. The footprint $\rR(i)$ is a set of indices that specifies which feature vectors are aggregated to construct the new feature $\yy_i$.

The function $\beta$ produces the feature vectors $\beta(\ve{x}_{j})$ that are aggregated by the adaptive weight vectors $\alpha(\ve{x}_i, \ve{x}_j)$. Possible instantiations of this function, along with feature transformation elements that surround self-attention operations in our architecture, are discussed later in this section.

The function $\alpha$ computes the weights $\alpha(\ve{x}_i, \ve{x}_j)$ that are used to combine the transformed features $\beta(\ve{x}_{j})$. To simplify exposition of different forms of self-attention, we decompose $\alpha$ as follows:
\begin{equation}
	\label{eq:pairwise-decomposition}
	\alpha(\ve{x}_i, \ve{x}_j) = \gamma(\delta(\ve{x}_{i},\ve{x}_{j})).
\end{equation}
The relation function $\delta$ outputs a single vector that represents the features $\xx_i$ and $\xx_j$. The function $\gamma$ then maps this vector into a vector that can be combined with $\beta(\ve{x}_{j})$ as shown in Eq.~\ref{eq:pairwise}.

The function $\gamma$ enables us to explore relations $\delta$ that produce vectors of varying dimensionality that need not match the dimensionality of $\beta(\ve{x}_{j})$. It also allows us to introduce additional trainable transformations into the construction of the weights $\alpha(\ve{x}_i, \ve{x}_j)$, making this construction more expressive. This function performs a linear mapping, followed by a nonlinearity, followed by another linear mapping; i.e., $\gamma$=\texttt{\{Linear$\rightarrow$ReLU$\rightarrow$Linear\}}. The output dimensionality of $\gamma$ does not need to match that of $\beta$ as attention weights can be shared across a group of channels.

We explore multiple forms for the relation function $\delta$:
\begin{description}
\item[Summation:] $\delta(\ve{x}_i, \ve{x}_j) = \varphi(\ve{x}_i)+\psi(\ve{x}_j)$
\item[Subtraction:] $\delta(\ve{x}_i, \ve{x}_j) = \varphi(\ve{x}_i)-\psi(\ve{x}_j)$
\item[Concatenation:] $\delta(\ve{x}_i, \ve{x}_j) = [\varphi(\ve{x}_i), \psi(\ve{x}_j)]$
\item[Hadamard product:] $\delta(\ve{x}_i, \ve{x}_j) = \varphi(\ve{x}_i)\odot\psi(\ve{x}_j)$
\item[Dot product:] $\delta(\ve{x}_i, \ve{x}_j) = \varphi(\ve{x}_i)^\top  \psi(\ve{x}_j)$
\end{description}

Here $\varphi$ and $\psi$ are trainable transformations such as linear mappings, and have matching output dimensionality. With summation, subtraction, and Hadamard product, the dimensionality of $\delta(\ve{x}_i, \ve{x}_j)$ is the same as the dimensionality of the transformation functions. With concatenation, the dimensionality of $\delta(\ve{x}_i, \ve{x}_j)$ will be doubled. With the dot product, the dimensionality of $\delta(\ve{x}_i, \ve{x}_j)$ is 1.

\paragraph{Position encoding.}
A distinguishing characteristic of pairwise attention is that feature vectors $\xx_j$ are processed independently and the weight computation $\alpha(\ve{x}_i, \ve{x}_j)$ cannot incorporate information from any location other than $i$ and $j$. To provide some spatial context to the model, we augment the feature maps with position information.
The position is encoded as follows. The horizontal and vertical coordinates along the feature map are first normalized to the range $[-1,1]$ in each dimension. These normalized two-dimensional coordinates are then passed through a trainable linear layer, which can map them to an appropriate range for each layer in the network. This linear mapping outputs a two-dimensional position feature $\pp_i$ for each location $i$ in the feature map. For each pair $(i,j)$ such that ${j \in \rR(i)}$, we encode the relative position information by calculating the difference $\pp_i-\pp_j$. The output of $\delta(\ve{x}_i, \ve{x}_j)$ is augmented by concatenating $[\pp_i-\pp_j]$ prior to the mapping $\gamma$.

\subsection{Patchwise Self-attention}
\label{sec:patchwise}

The other type of self-attention we explore is referred to as \emph{patchwise} and has the following form:
\begin{equation}
	\label{eq:patchwise}
	\ve{y}_{i} = \sum_{j \in \rR(i)} \alpha(\xx_{\rR(i)})_j \odot \beta(\ve{x}_{j}),
\end{equation}
where $\xx_{\rR(i)}$ is the patch of feature vectors in the footprint $\rR(i)$. $\alpha(\xx_{\rR(i)})$ is a tensor of the same spatial dimensionality as the patch $\xx_{\rR(i)}$. $\alpha(\xx_{\rR(i)})_j$ is the vector at location $j$ in this tensor, corresponding spatially to the vector $\xx_j$ in $\xx_{\rR(i)}$.

In patchwise self-attention, we allow the construction of the weight vector that is applied to $\beta(\ve{x}_{j})$ to refer to and incorporate information from all feature vectors in the footprint $\rR(i)$. Note that, unlike pairwise self-attention, patchwise self-attention is no longer a set operation with respect to the features $\xx_j$. It is not permutation-invariant or cardinality-invariant: the weight computation $\alpha(\xx_{\rR(i)})$ can index the feature vectors $\xx_j$ individually, by location, and can intermix information from feature vectors from different locations within the footprint. Patchwise self-attention is thus strictly more powerful than convolution.

We decompose $\alpha(\xx_{\rR(i)})$ as follows:
\begin{equation}
	\label{eq:patchwise-decomposition}
	\alpha(\xx_{\rR(i)}) = \gamma(\delta(\xx_{\rR(i)})).
\end{equation}
The function $\gamma$ maps a vector produced by $\delta(\xx_{\rR(i)})$ to a tensor of appropriate dimensionality. This tensor comprises weight vectors for all locations $j$.
The function $\delta$ combines the feature vectors $\xx_j$ from the patch $\xx_{\rR(i)}$. We explore the following forms for this combination:
\begin{description}
\item[Star-product:]
$\delta(\xx_{\rR(i)}) = [\varphi(\ve{x}_i)^\top \psi(\ve{x}_j)]_{\forall j \in \rR(i)}$
\item[Clique-product:]
$\delta(\xx_{\rR(i)}) = [\varphi(\ve{x}_j)^\top \psi(\ve{x}_k)]_{\forall j,k \in \rR(i)}$
\item[Concatenation:]
$\delta(\xx_{\rR(i)}) = [\varphi(\ve{x}_i), [ \psi(\ve{x}_j)]_{\forall j \in \rR(i)}]$
\end{description}

\newcommand{\blocka}[3]{\multirow{2}{*}{\(\left[\begin{array}{c}\text{3$\times$3, #1-d sa}\\[-.1em] \text{#2-d linear} \end{array}\right]\)$\times$#3}}
\newcommand{\blockb}[3]{\multirow{2}{*}{\(\left[\begin{array}{c}\text{5$\times$5, #1-d sa}\\[-.1em] \text{#2-d linear} \end{array}\right]\)$\times$#3}}
\newcommand{\blockc}[3]{\multirow{2}{*}{\(\left[\begin{array}{c}\text{7$\times$7, #1-d sa}\\[-.1em] \text{#2-d linear} \end{array}\right]\)$\times$#3}}

\renewcommand\arraystretch{1.1}
\setlength{\tabcolsep}{4pt}
\begin{table*}[th!]
\centering
		\resizebox{0.9\linewidth}{!}{
			\begin{tabular}{c|c|c|c|c}
				\toprule[1pt]
				Layers & Output Size & SAN10 & SAN15 & SAN19 \\
				\hline
				Input & 224$\times$224$\times$3 & \multicolumn{3}{c}{64-d linear}\\
				\hline
				{Transition} & {112$\times$112$\times$64} & \multicolumn{3}{c}{2$\times$2, stride 2 max pool $\rightarrow$ 64-d linear} \\\cline{3-5}
				\hline
				\multirow{2}{*}{SA Block} & \multirow{2}{*}{112$\times$112$\times$64} & \blocka{16}{64}{2} & \blocka{16}{64}{3} & \blocka{16}{64}{3} \\
				&  &  &  &\\
				\hline
				{Transition} & {56$\times$56$\times$256} & \multicolumn{3}{c}{2$\times$2, stride 2 max pool $\rightarrow$ 256-d linear} \\\cline{3-5}
				\hline
				\multirow{2}{*}{SA Block} & \multirow{2}{*}{56$\times$56$\times$256} & \blockc{64}{256}{1} & \blockc{64}{256}{2} & \blockc{64}{256}{3} \\
				&  &  &  &\\
				\hline
				{Transition} & {28$\times$28$\times$512} & \multicolumn{3}{c}{2$\times$2, stride 2 max pool $\rightarrow$ 512-d linear} \\\cline{3-5}
				\hline
				\multirow{2}{*}{SA Block} & \multirow{2}{*}{28$\times$28$\times$512} & \blockc{128}{512}{2} & \blockc{128}{512}{3} & \blockc{128}{512}{4} \\
				&  &  &  &\\
				\hline
				{Transition} & {14$\times$14$\times$1024} & \multicolumn{3}{c}{2$\times$2, stride 2 max pool $\rightarrow$ 1024-d linear} \\\cline{3-5}
				\hline
				\multirow{2}{*}{SA Block} & \multirow{2}{*}{14$\times$14$\times$1024} & \blockc{256}{1024}{4} & \blockc{256}{1024}{5} & \blockc{256}{1024}{6} \\
				&  &  &  &\\
				\hline
				{Transition} & {7$\times$7$\times$2048} & \multicolumn{3}{c}{2$\times$2, stride 2 max pool $\rightarrow$ 2048-d linear} \\\cline{3-5}
				\hline
				\multirow{2}{*}{SA Block} & \multirow{2}{*}{7$\times$7$\times$2048} & \blockc{512}{2048}{1} & \blockc{512}{2048}{2} & \blockc{512}{2048}{3} \\
				&  &  &  &\\
				\hline
				Classification & 1$\times$1$\times$1000  & \multicolumn{3}{c}{global average pool $\rightarrow$ 1000-d linear $\rightarrow$ softmax} \\
				\bottomrule[1pt]
			\end{tabular}
		}
		\vspace{1mm}
	\caption{Self-attention networks for image recognition. `C-d linear' means that the output dimensionality of the linear layer is `C'. `C-d sa' stands for a self-attention operation with output dimensionality `C'. SAN10, SAN15, and SAN19 are in rough correspondence with ResNet26, ResNet38, and ResNet50, respectively. The number X in SANX refers to the number of self-attention blocks. Our architectures are based fully on self-attention.}
	\label{tab:arch-imagenet}
	\vspace{-3mm}
\end{table*}

\subsection{Self-attention Block}
\label{sec:sab-detail}

\begin{figure}[ht]
	\centering
		\includegraphics[width=0.65\linewidth]{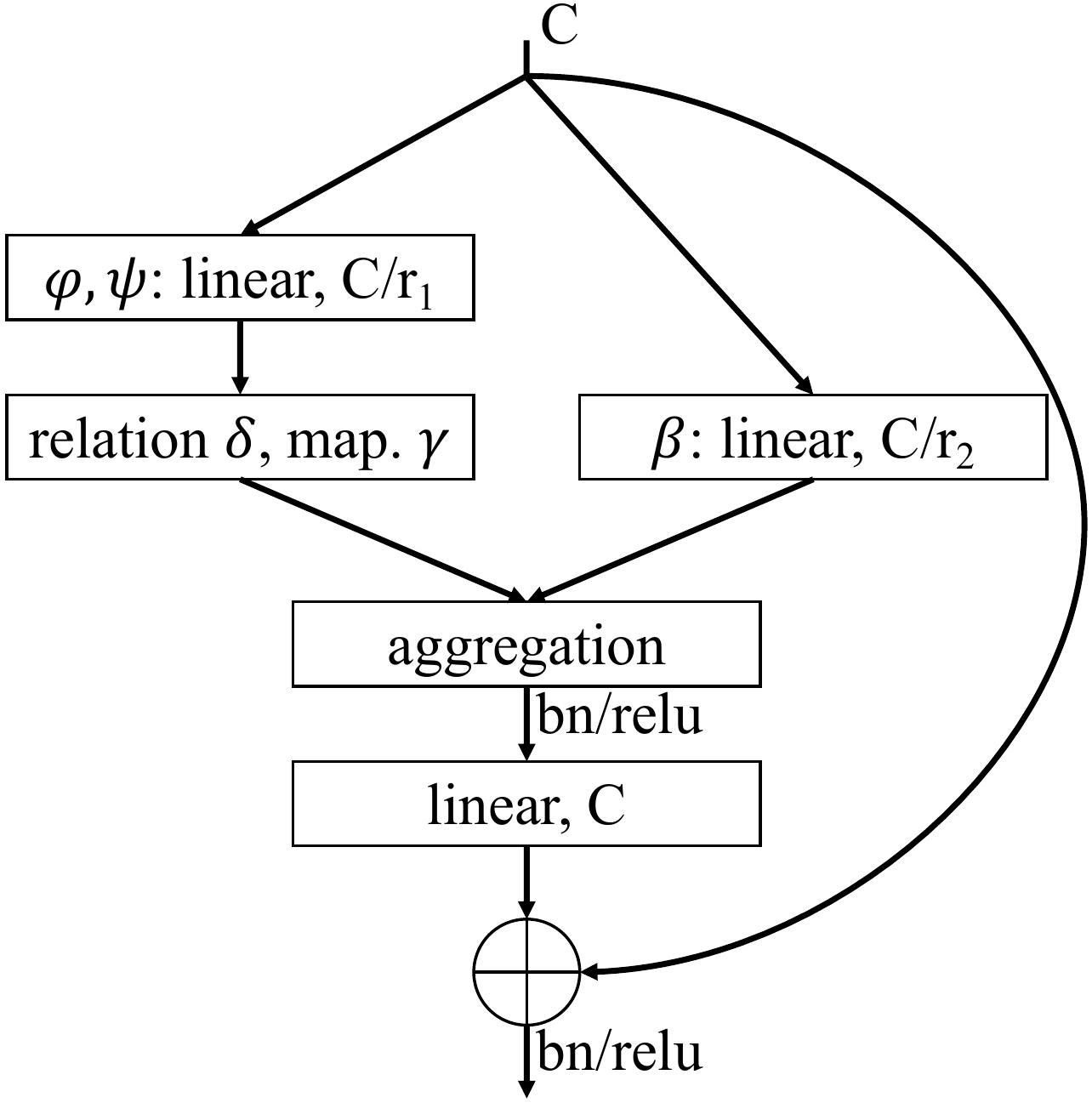}
	\caption{Our self-attention block. $C$ is the channel dimensionality. The left stream evaluates the attention weights $\alpha$, the right stream transforms the features via a linear mapping $\beta$. Both streams reduce the channel dimensionality for efficient processing. The outputs of the streams are aggregated via a Hadamard product and the dimensionality is subsequently expanded back to $C$.}
	\label{fig:sab}
	\vspace{-0.1in}
\end{figure}

The self-attention operations described in Sections~\ref{sec:pairwise} and~\ref{sec:patchwise} can be used to construct residual blocks~\cite{he2016deep} that perform both feature aggregation and feature transformation. Our self-attention block is illustrated in Figure~\ref{fig:sab}. The input feature tensor (channel dimensionality $C$) is passed through two processing streams. The left stream evaluates the attention weights $\alpha$ by computing the function $\delta$ (via the mappings $\varphi$ and $\psi$) and a subsequent mapping $\gamma$. The right stream applies a linear transformation $\beta$ that transforms the input features and reduces their dimensionality for efficient processing. The outputs of the two streams are then aggregated via a Hadamard product. The combined features are passed through a normalization and an elementwise nonlinearity, and are processed by a final linear layer that expands their dimensionality back to $C$.

\subsection{Network Architectures}
\label{sec:san}

Our network architectures generally follow residual networks, which we will use as baselines~\cite{he2016deep}. Table~\ref{tab:arch-imagenet} presents three architectures obtained by stacking self-attention blocks at different resolutions. These architectures~-- SAN10, SAN15, and SAN19~-- are in rough correspondence with ResNet26, ResNet38, and ResNet50. The number X in SANX refers to the number of self-attention blocks. Our architectures are based fully on self-attention.

\paragraph{Backbone.}
The backbone of SAN has five stages, each with different spatial resolution, yielding a resolution reduction factor of 32. Each stage comprises multiple self-attention blocks. Consecutive stages are bridged by transition layers that reduce spatial resolution and expand channel dimensionality. The output of the last stage is processed by a classification layer that comprises global average pooling, a linear layer, and a softmax.

\paragraph{Transition.}
Transition layers reduce spatial resolution, thus reducing the computational burden and expanding receptive field. The transition comprises a batch normalization layer, a ReLU~\cite{nair2010rectified}, $2\timess 2$ max pooling with stride 2, and a linear mapping that expands channel dimensionality.

\paragraph{Footprint of self-attention.}
The local footprint $\rR(i)$ controls the amount of context gathered by a self-attention operator from the preceding feature layer. We set the footprint size to $7\timess 7$ for the last four stages of SAN. The footprint is set to $3 \timess 3$ in the first stage due to the high resolution of that stage and the consequent memory consumption. Note that increasing the footprint size has no impact on the number of parameters in pairwise self-attention. We will study the effect of footprint size on accuracy, capacity, and FLOPs in Section~\ref{sec:abla}.

\paragraph{Instantiations.}
The number of self-attention blocks in each stage can be adjusted to obtain networks with different capacities.
In the networks presented in Table~\ref{tab:arch-imagenet}, the number of self-attention blocks used in the last four stages is the same as the number of residual blocks in ResNet26, ResNet38, and ResNet50, respectively.

\section{Comparison}
In this section, we relate the family of self-attention operators presented in Section~\ref{sec:method} to other constructions, including convolution~\cite{LeCun1989} and scalar attention~\cite{vaswani2017attention,wang2017non,ramachandran2019stand,hu2019local}. Table~\ref{tab:properties} summarizes some differences between the constructions. These are discussed in more detail below.

\begin{table}[h]
\centering
\footnotesize
\setlength{\tabcolsep}{2.0pt}
\begin{tabular}{l|c|c}
\toprule[1pt]
Operation & Content adaptive & Channel adaptive  \\
\hline
Convolution~\cite{LeCun1989} & \xmark & \cmark \\
\hline
Scalar attention~\cite{vaswani2017attention,wang2017non,ramachandran2019stand,hu2019local} & \cmark & \xmark \\
\hline
Vector attention (ours) & \cmark & \cmark \\
\bottomrule[1pt]
\end{tabular}
\vspace{1mm}
\caption{The convolution does not adapt to the content of the image. Scalar attention produces scalar weights that do not vary along the channel dimension. Our operators efficiently compute attention weights that adapt across both spatial dimensions and channels.}
\label{tab:properties}
\vspace{-3mm}
\end{table}

\paragraph{Convolution.}
The regular convolution operator has fixed kernel weights that are independent of the content of the image. It does not adapt to the input content. The kernel weights can vary across channels.

\paragraph{Scalar attention.}
Scalar attention, as used in the transformer~\cite{vaswani2017attention} and related constructions in computer vision~\cite{wang2017non,ramachandran2019stand,hu2019local}, typically has the following form:
\begin{equation}\label{eq:transformer}
\ve{y}_{i} = \sum_{j \in \rR(i)} \big(\varphi(\ve{x}_i)^\top \psi(\ve{x}_j)\big) \beta(\ve{x}_{j})
\end{equation}
(A softmax and other forms of normalization can be added.)
Unlike the convolution, the aggregation weights can vary across different locations, depending on the content of the image. On the other hand, the weight $\varphi(\ve{x}_i)^\top \psi(\ve{x}_j)$ is a \emph{scalar} that is shared across all channels. (Hu et al.~\cite{hu2019local} explored alternatives to the dot product, but these alternatives operated on scalar weights that were likewise shared across channels.) This construction does not adapt the attention weights at different channels. Although this can be mitigated to some extent by introducing multiple heads~\cite{vaswani2017attention}, the number of heads is a small constant and scalar weights are shared by all channels within a head.

\begin{table*}
	\centering
	\begin{tabular}{l|c|c|c|c|c|c|c|c|c|c|c|c}
		\toprule[1pt]
		\multirow{2}{*}{Method} & \multicolumn{4}{c|}{ResNet26 vs.\ SAN10} & \multicolumn{4}{c|}{ResNet38 vs.\ SAN15} & \multicolumn{4}{c}{ResNet50 vs.\ SAN19} \\
		\cline{2-13}
		& top-1 & top-5 & Params & Flops & top-1 & top-5 & Params & Flops & top-1 & top-5 & Params & Flops \\
		\hline
		Convolutional & 73.6 & 91.7 & 13.7M & 2.4G & 76.0 & 93.0 & 19.6M & 3.2G & 76.9 & 93.5 & 25.6M & 4.1G \\
		SAN, pairwise & 74.9 & 92.1 & 10.5M & 2.2G & 76.6 & 93.1 & 14.1M & 3.0G & 76.9 & 93.4 & 17.6M & 3.8G \\
		SAN, patchwise & 77.1 & 93.5 & 11.8M & 1.9G & 78.0 & 93.9 & 16.2M & 2.6G & 78.2 & 93.9 & 20.5M & 3.3G \\
		\bottomrule[1pt]
	\end{tabular}
	\vspace{1mm}
	\caption{Comparison of self-attention networks and convolutional residual networks on ImageNet classification. Single-crop testing on the val-original set.}
	\label{tab:imagenet}
\vspace{-3mm}
\end{table*}

\paragraph{Vector attention (ours).}
The operators presented in Section~\ref{sec:method} subsume scalar attention and generalize it in important ways. First, within the pairwise attention family, the relation function $\delta$ can produce \emph{vector} output. This is the case for the summation, subtraction, Hadamard, and concatenation forms. This vector can then be further processed and mapped to the right dimensionality by $\gamma$, which can also take position encoding channels as input. The mapping $\gamma$ produces a vector that has compatible dimensionality to the transformed features $\beta$. This gives the construction significant flexibility in accommodating different relation functions and auxiliary inputs, expressive power due to multiple linear mappings and nonlinearities along the computation graph, ability to produce attention weights that vary along both spatial and channel dimensions, and computational efficiency due to the ability to reduce dimensionality by the mappings $\gamma$ and $\beta$.

The patchwise family of operators generalizes convolution while retaining parameter and FLOP efficiency. This family of operators produces weight vectors for all positions along a feature map that also vary along the channel dimension. The weight vectors are informed by the entirety of the footprint of the operator.

%% file: tex/experiments.tex
We conduct experiments on ImageNet classification~\cite{Russakovsky2015}. The dataset contains 1.28 million training images and 50K validation images from 1000 different classes. For comparisons of self-attention networks with convolutional networks such as ResNet, we train on the original training set and report accuracy (single center crop) on the original validation set (referred to as `val-original'). For controlled experiments and ablation studies on self-attention networks, we split a separate validation set out of the original training set by randomly sampling 50 images from the training set for each category: this is referred to as `val-split'. This ensures that architectural and hyperparameter choices are not made on the same set that is used for comparisons with external baselines.

\subsection{Implementation}
\label{sec:implementation}

We train all models from scratch for 100 epochs. We use the cosine learning rate schedule with base learning rate 0.1~\cite{loshchilov2017sgdr}. We apply standard data augmentation on ImageNet, including random cropping to $224\timess 224$ patches~\cite{szegedy2015going}, random horizontal flipping, and normalization. We use synchronous SGD with minibatch size 256 on 8 GPUs. We use label smoothing regularization with coefficient 0.1~\cite{szegedy2016rethinking}. Momentum and weight decay are set to 0.9 and 1e-4, respectively~\cite{he2016deep,xie2017aggregated,goyal2017accurate}.

Our convolutional network baselines are ResNet26, ResNet38, and ResNet50~\cite{he2016deep}. ResNet38 and ResNet26 are constructed by taking ResNet50 as the starting point and removing one or two residual blocks from each stage, respectively. For self-attention blocks, we use ${r_1 = 16}$ and ${r_2 = 4}$ by default (see Figure~\ref{fig:sab} for notation). The number of channels sharing the same attention weight is set to 8.

\subsection{Comparison to Convolutional Networks}
\label{sec:main-results}

Table~\ref{tab:imagenet} reports the results of the main comparison of the presented self-attention networks to convolutional counterparts. For pairwise self-attention, we use the subtraction relation. For patchwise self-attention, we use concatenation. These decisions are based on the controlled experiments reported in Section~\ref{sec:abla}. The pairwise models match or outperform the convolutional baselines, with similar or lower parameter and FLOP budgets. The patchwise models perform even better. For example, the patchwise SAN10 outperforms not only ResNet26 but also ResNet38, with a 40\% lower parameter count and a 41\% lower FLOP count versus the latter. Likewise, the patchwise SAN15 outperforms not only ResNet38 but also ResNet50 (78\% top-1 accuracy for SAN15 versus 76\% for ResNet38 and 76.9\% for ResNet50), with a 37\% lower parameter count and a 37\% lower FLOP count versus the latter.

\subsection{Controlled Experiments}
\label{sec:abla}

\paragraph{Relation function.}
Table~\ref{tab:relation} reports the results of a controlled comparison of different relation functions on the val-split set. For pairwise self-attention, summation, subtraction, and Hadamard product achieve similar accuracy. These relation functions outperform concatenation and dot product. In particular, these experiments indicate that vector self-attention outperforms scalar self-attention. For patchwise self-attention, concatenation achieves slightly higher accuracy than star-product and clique-product.

\begin{table}[h]
	\centering
	\resizebox{1.0\linewidth}{!}{
		\begin{tabular}{l|l|c|c|c|c}
			\toprule[1pt]
			\multicolumn{2}{c|}{Method} & top-1 & top-5 & Params & Flops \\
			\hline
			\multicolumn{2}{c|}{Conv.-ResNet26} & 76.0  & 92.8 & 13.7M & 2.4G \\
			\hline
			\multirow{5}{*}{SAN10-pair.} & summation & 77.4 & 93.3 & 10.5M & 2.2G \\
			& subtraction & 77.4 & 93.3 & 10.5M & 2.2G \\
			& concatenate & 76.4 & 92.6 & 10.6M & 2.5G \\
			& Had. product & 77.4 & 93.4 & 10.5M & 2.2G \\
			& dot product & 77.0 & 93.0 & 10.5M & 1.8G \\
			\hline
			\multirow{3}{*}{SAN10-patch.} & star-product & 78.7 & 94.0 & 10.9M & 1.7G \\
			& clique-product & 79.1 & 94.2 & 11.5M & 1.9G \\
			& concatenation & 79.3 & 94.2 & 11.8M & 1.9G \\
			\bottomrule[1pt]
	\end{tabular}}
	\vspace{1mm}
	\caption{Controlled comparison of different relation functions on the val-split set.}
	\label{tab:relation}
	\vspace{-3mm}
\end{table}

We also attempted a controlled comparison with the self-attention configuration of Ramachandran et al.~\cite{ramachandran2019stand}. Unfortunately, their implementation has not been released at the time of writing, and there are many subtle differences that can impact results, from the configuration of the input stem, to positional encoding, to architectural hyperparameters, to data augmentation and the training schedule. We attempted to control for extraneous differences as much as possible by using the same overall network architecture (SAN10) and training setup (Section~\ref{sec:implementation}). Within this framework, we reproduced the self-attention block of Ramachandran et al.\ as closely as possible. In particular, we used their grouped dot-product attention, added position information, and set $r_1$ and $r_2$ (the bottleneck dimension reduction factor) to 4. This yielded top-1 accuracy of 71.7\% and top-5 accuracy of 89.9\%, lower than our self-attention configurations with the same setup and lower than the results reported by Ramachandran et al. (The number of parameters is 13.9M, the number of FLOPs is 2.3G.) Considered in conjunction with our controlled experiments, this appears to support the conclusion that vector self-attention is a useful building block for self-attention networks in computer vision. Our results also indicate that patchwise self-attention may be particularly powerful and merits further study. Finally, the difficulties in reproducing results reported in related work highlight the importance of timely release of reference implementations. We will release our full implementation and experimental setup open-source to facilitate comparison and assist future work in this area.

\paragraph{Mapping function.}
We conduct an ablation study on the number of linear layers in the attention mapping function $\gamma$. The results are listed in Table~\ref{tab:mapping}. For pairwise models, using two linear layers yields the highest accuracy. For patchwise models, different settings yield similar accuracy. Using only one linear layer for attention mapping incurs significant memory and computation costs in the patchwise setting. Multiple layers enable the introduction of bottlenecks that reduce dimensionality and thus reduce memory and computation costs. Considering all the factors, we use two linear layers (the intermediate setting in Table~\ref{tab:mapping}) as our default for all models.

\begin{table}[h]
	\centering
	\resizebox{1.0\linewidth}{!}{
		\begin{tabular}{l|c|c|c|c|c}
			\toprule[1pt]
			\multicolumn{2}{c|}{Method} & top-1 & top-5 & Params & Flops \\
			\hline
			\multicolumn{2}{c|}{Conv.-ResNet26} & 76.0 & 92.8 & 13.7M & 2.4G \\
			\hline
			\multirow{3}{*}{SAN10-pair.} & \texttt{L} & 75.8 & 92.3 & 10.5M & 1.8G \\
			& \texttt{L$\rightarrow$R$\rightarrow$L} & 77.4 & 93.3 & 10.5M & 2.2G \\
			& \texttt{L$\rightarrow$R$\rightarrow$L$\rightarrow$R$\rightarrow$L} & 77.0 & 93.0 & 10.6M & 2.5G \\
			\hline
			\multirow{3}{*}{SAN10-patch.} & \texttt{L} & 79.3 & 94.2 & 53.5M & 9.5G \\
			& \texttt{L$\rightarrow$R$\rightarrow$L} & 79.3 & 94.2 & 11.8M & 1.9G \\
			& \texttt{L$\rightarrow$R$\rightarrow$L$\rightarrow$R$\rightarrow$L} & 79.5 & 94.3 & 12.7M & 2.0G \\
			\bottomrule[1pt]
	\end{tabular}}
	\vspace{1mm}
	\caption{Controlled comparison of different mapping functions on the val-split set. \texttt{L} and \texttt{R} denote \texttt{Linear} and \texttt{ReLU} layers, respectively.}
	\label{tab:mapping}
	\vspace{-3mm}
\end{table}

\paragraph{Transformation functions.}
We now evaluate whether the use of three distinct transformation functions ($\varphi$, $\psi$, and $\beta$) is helpful. The results are reported in Table~\ref{tab:transformation}. Using three distinct learnable transformations is generally the best choice. An additional advantage is that a distinct $\beta$ transformation enables the use of different bottleneck dimension reduction factors $r_1$ and $r_2$, which can be used to lower FLOP consumption. For $\varphi=\psi=\beta$, we set $r_1 = r_2 = 4$, which yields comparable accuracy to $\varphi=\psi\ne\beta$ but at higher FLOP counts.

\begin{table}[h]
	\centering
	\resizebox{1.0\linewidth}{!}{
		\begin{tabular}{l|c|c|c|c|c}
			\toprule[1pt]
			\multicolumn{2}{c|}{Method} & top-1 & top-5 & Params & Flops \\
			\hline
			\multicolumn{2}{c|}{Conv.-ResNet26} & 76.0 & 92.8 & 13.7M & 2.4G \\
			\hline
			\multirow{3}{*}{SAN10-pair.} & $\varphi=\psi=\beta$ & 76.5 & 92.8 & 9.5M & 3.0G \\
			& $\varphi=\psi\ne\beta$ & 76.3 & 92.6 & 10.0M & 2.1G \\
			& $\varphi\ne\psi\ne\beta$ & 77.4 & 93.3 & 10.5M & 2.2G \\
			\hline
			\multirow{3}{*}{SAN10-patch.} & $\varphi=\psi=\beta$ & 78.9 & 94.1 & 13.4M & 2.2G \\
			& $\varphi=\psi\ne\beta$ & 79.0 & 94.0 & 11.3M & 1.8G \\
			& $\varphi\ne\psi\ne\beta$ & 79.3 & 94.2 & 11.8M & 1.9G \\
			\bottomrule[1pt]
	\end{tabular}}
	\vspace{1mm}
	\caption{Controlled evaluation of the use of distinct transformation functions.}
	\label{tab:transformation}
	\vspace{-3mm}
\end{table}

\paragraph{Footprint size.}
We now assess the impact of the size of the footprint $\rR(i)$ of the self-attention operator. The results are reported in Table~\ref{tab:footprint}. In convolutional networks, larger footprint sizes incur significant memory and computation costs. In self-attention networks, the accuracy initially increases with footprint size and then saturates. For pairwise self-attention, increasing the footprint size has no impact on the number of parameters. Taking all factors into account, we set the footprint size to 7$\times$7 as our default for all models.

\begin{table}[h]
	\centering
	\resizebox{1.0\linewidth}{!}{
	\begin{tabular}{l|c|c|c|c|c}
		\toprule[1pt]
		\multicolumn{2}{c|}{Method} & top-1 & top-5 & Params & Flops \\
		\hline
		\multirow{3}{*}{Conv.-ResNet26} & 3$\times$3 & 76.0  & 92.8 & 13.7M & 2.4G \\
		& 5$\times$5 & 77.4 & 93.6 & 22.7M & 4.0G \\
		& 7$\times$7 & 77.9 & 93.7 & 36.1M & 6.5G \\
		\hline
		\multirow{5}{*}{SAN10-pair.} & 3$\times$3 & 75.3 & 92.0 & 10.5M & 1.7G \\
		& 5$\times$5 & 76.6 & 92.9 & 10.5M & 1.9G \\
		& 7$\times$7 & 77.4 & 93.3 & 10.5M & 2.2G \\
		& 9$\times$9 & 77.8 & 93.5 & 10.5M & 2.5G \\
		& 11$\times$11 & 77.6 & 93.3 & 10.5M & 3.0G \\
		\hline
		\multirow{5}{*}{SAN10-patch.} & 3$\times$3 & 77.4  & 93.4 & 10.7M & 1.6G \\
		& 5$\times$5 & 78.7 & 94.0 & 11.2M & 1.7G \\
		& 7$\times$7 & 79.3 & 94.2 & 11.8M & 1.9G \\
		& 9$\times$9 & 79.3 & 94.1 & 12.7M & 2.1G \\
		& 11$\times$11 & 79.4 & 94.1 & 13.8M & 2.3G \\
		\bottomrule[1pt]
	\end{tabular}}
	\vspace{1mm}
	\caption{Controlled assessment of the impact of footprint size.}
	\label{tab:footprint}
	\vspace{-3mm}
\end{table}

\paragraph{Position encoding.}
Finally, we evaluate the importance of position encoding in pairwise self-attention. The results are reported in Table~\ref{tab:position}. Position encoding has a significant effect. Without position encoding, top-1 accuracy drops by 5 percentage points. Absolute position encoding~\cite{liu2018intriguing} is better than none, but accuracy is still low. Relative position encoding, as described in Section~\ref{sec:pairwise}, is much more effective.

\begin{table}[h]
	\centering
	\begin{tabular}{l|l|c|c|c|c}
		\toprule[1pt]
		\multicolumn{2}{c|}{Method} & top-1 & top-5 & Params & Flops \\
		\hline
		\multicolumn{2}{c|}{Conv.-ResNet26} & 76.0  & 92.8 & 13.7M & 2.4G \\
		\hline
		\multirow{3}{*}{SAN10-pair.} & none & 72.3 & 90.3 & 10.5M & 2.1G \\
		& absolute & 74.7 & 91.7 & 10.5M & 2.2G \\
		& relative & 77.4 & 93.3 & 10.5M & 2.2G \\
		\bottomrule[1pt]
	\end{tabular}
	\vspace{1mm}
	\caption{The importance of position encoding in pairwise self-attention.}
	\label{tab:position}
	\vspace{-3mm}
\end{table}

\begin{table*}
	\centering
	\resizebox{1.0\linewidth}{!}{
	\begin{tabular}{l|c|c|c|c|c|c|c|c|c|c}
		\toprule[1pt]
		\multirow{2}{*}{Method} & \multicolumn{2}{c|}{no rotation} & \multicolumn{2}{c|}{clockwise 90$^\circ$} & \multicolumn{2}{c|}{clockwise 180$^\circ$} & \multicolumn{2}{c|}{clockwise 270$^\circ$} & \multicolumn{2}{c}{upside-down} \\
		\cline{2-11}
		& top-1 & top-5 & top-1 & top-5 & top-1 & top-5 & top-1 & top-5 & top-1 & top-5 \\
		\hline
		ResNet26 & 73.6 & 91.7 & 49.1\footnotesize{(24.5)} & 72.7\footnotesize{(19.0)} & 50.6\footnotesize{(23.0)} & 75.4\footnotesize{(16.3)} & 49.2\footnotesize{(24.4)} & 72.8\footnotesize{(18.9)} & 50.5\footnotesize{(23.1)} & 75.4\footnotesize{(16.3)} \\
		SAN10-pair. & 74.9 & 92.1 & 51.8\footnotesize{(\textbf{23.1})} & 74.6\footnotesize{(\textbf{17.5})} & 54.7\footnotesize{(\textbf{20.2})} & 78.5\footnotesize{(\textbf{13.6})} & 51.7\footnotesize{(23.2)} & 74.5\footnotesize{(\textbf{17.6})} & 54.7\footnotesize{(\textbf{20.2})} & 78.5\footnotesize{(\textbf{13.6})} \\
		SAN10-patch. & 77.1 & 93.5 & 53.1\footnotesize{(24.0)} & 75.7\footnotesize{(17.8)} & 54.6\footnotesize{(22.5)} & 78.4\footnotesize{(15.1)} & 53.3\footnotesize{(23.8)} & 76.0\footnotesize{(17.5)} & 54.7\footnotesize{(22.4)} & 78.3\footnotesize{(15.2)} \\
		\hline
		ResNet38 & 76.0 & 93.0 & 51.2\footnotesize{(24.8)} & 74.2\footnotesize{(18.8)} & 52.2\footnotesize{(23.8)} & 76.9\footnotesize{(16.1)} & 51.6\footnotesize{(24.4)} & 74.6\footnotesize{(18.4)} & 52.2\footnotesize{(23.8)} & 76.8\footnotesize{(16.2)} \\
		SAN15-pair. & 76.6 & 93.1 & 54.5\footnotesize{(\textbf{22.1})} & 77.1\footnotesize{(\textbf{16.0})} & 57.9\footnotesize{(\textbf{18.7})} & 80.8\footnotesize{(\textbf{12.3})} & 54.8\footnotesize{(\textbf{21.8})} & 77.0\footnotesize{(\textbf{16.1})} & 58.0\footnotesize{(\textbf{18.6})} & 80.8\footnotesize{(\textbf{12.3})} \\
		SAN15-patch. & 78.0 & 93.9 & 53.7\footnotesize{(24.5)} & 76.1\footnotesize{(17.8)} & 56.0\footnotesize{(22.2)} & 79.5\footnotesize{(14.4)} & 53.9\footnotesize{(24.3)} & 76.2\footnotesize{(17.7)} & 56.0\footnotesize{(22.2)} & 79.4\footnotesize{(14.5)} \\
		\hline
		ResNet50 & 76.9 & 93.5 & 52.6\footnotesize{(24.3)} & 75.3\footnotesize{(18.2)} & 52.9\footnotesize{(24.0)} & 77.4\footnotesize{(16.2)} & 52.6\footnotesize{(24.3)} & 75.5\footnotesize{(18.0)} & 53.0\footnotesize{(23.9)} & 77.3\footnotesize{(16.2)} \\
		SAN19-pair. & 76.9 & 93.4 & 54.7\footnotesize{(\textbf{22.2})} & 77.1\footnotesize{(\textbf{16.3})} & 58.0\footnotesize{(\textbf{18.9})} & 80.4\footnotesize{(\textbf{13.0})} & 55.0\footnotesize{(\textbf{\textbf{21.9}})} & 77.1\footnotesize{(\textbf{16.3})} & 57.9\footnotesize{(\textbf{19.0})} & 80.4\footnotesize{(\textbf{13.0})} \\
		SAN19-patch. & 78.2 & 93.9 & 54.2\footnotesize{(24.0)} & 76.3\footnotesize{(17.6)} & 56.2\footnotesize{(22.0)} & 79.5\footnotesize{(14.4)} & 54.1\footnotesize{(24.1)} & 76.4\footnotesize{(17.5)} & 56.3\footnotesize{(21.9)} & 79.5\footnotesize{(14.4)} \\
		\bottomrule[1pt]
	\end{tabular}}
	\vspace{1mm}
	\caption{Robustness of trained networks to rotation and flipping of images at test time. Zero-shot testing on the val-original set. Numbers in the brackets show the relative performance drop compared to testing on original images with no manipulation (lower is better). Pairwise self-attention models are less vulnerable than convolutional networks or patchwise self-attention.}
	\label{tab:rotation}
	\vspace{-3mm}
\end{table*}

\subsection{Robustness}

We now conduct two experiments that probe the robustness of the representations learned by self-attention networks, as compared to convolutional baselines.

\paragraph{Zero-shot generalization to rotated images.}
The first experiment tests trained networks on rotated and flipped images. In this experiment, ImageNet images from the val-original set are rotated and flipped in one of four ways: clockwise 90$^\circ$, clockwise 180$^\circ$, clockwise 270$^\circ$, and upside-down flip about the horizontal axis. This is zero-shot testing: such manipulations were not performed at training time.

The results are reported in Table~\ref{tab:rotation}. Our hypothesis was that pairwise self-attention models will be more robust to this kind of manipulation than convolutional networks (or patchwise self-attention), given that pairwise self-attention is fundamentally a set operator. Indeed, we see that pairwise self-attention models are less vulnerable than convolutional or patchwise self-attention networks, although all networks suffer from the domain shift. For example, when images are rotated by 180$^\circ$, the performance of pairwise SAN19 drops by 18.9 percentage points, which is 5.1 percentage points lower than the drop suffered by ResNet50. The pairwise SAN10 model achieves 54.7\% top-1 accuracy in this regime, which is higher than the accuracy of the much larger ResNet50 (52.9\%).

\paragraph{Robustness to adversarial attacks.}
Next, we evaluate the robustness of trained networks to adversarial attacks. We subject the trained models to white-box targeted PGD attacks~\cite{madry2018towards}. Hyperparameters of the attacks include the maximal per-pixel perturbation $\epsilon$ (under the $L_\infty$ norm), attack step size $\rho$, and the number of attack iterations $n$. We test with two sets of hyperparameters: \{$\epsilon$, $\rho$, $n$\} set to \{8, 4, 2\} and \{8, 2, 4\}, respectively. The results are reported in Table~\ref{tab:attack}.

The results indicate that self-attention models are much more robust than convolutional networks. For example, with 4 attack iterations, the attack success rate for ResNet50 is 82.5\% and top-1 accuracy drops to 11.8\%. For the corresponding pairwise and patchwise SAN models, the attack success rate is much lower, at 63.7\% and 62.0\%, respectively, and the models' accuracy is roughly 2x higher, at 21.8\% and 24.8\%, respectively. For the ResNet26 baseline, 4 attack iterations essentially destroy the model, with a top-1 accuracy of 1\%. In comparison, the top-1 accuracy of the patchwise SAN model is roughly 10x higher at 9.6\%. (A random guess baseline would exhibit a top-1 accuracy of 0.1\%.)

\begin{table}[h]
	\centering
	\begin{tabular}{l|c|c|c|c|c}
		\toprule[1pt]
		\multirow{2}{*}{Method} & clean & \multicolumn{2}{c|}{attack $n=2$} & \multicolumn{2}{c}{attack $n=4$} \\
		\cline{2-6}
		& top-1 & s.\ rate & top-1 & s.\ rate & top-1 \\
		\hline
		ResNet26 & 73.6 & 49.0 & 26.6 & 98.2 & 1.0 \\
		SAN10-pair. & 74.9 & 32.8 & 35.3 & 90.1 & 5.3 \\
		SAN10-patch. & 77.1 & 24.5 & 46.4 & 85.8 & 9.6 \\
		\hline
		ResNet38 & 76.0 & 32.7 & 39.2 & 94.1 & 3.8 \\
		SAN15-pair. & 76.6 & 15.5 & 47.3 & 67.5 & 19.6 \\
		SAN15-patch. & 78.0 & 13.1 & 54.8 & 65.6 & 22.9 \\
		\hline
		ResNet50 & 76.9 & 19.5 & 49.3 & 82.5 & 11.8 \\
		SAN19-pair. & 76.9 & 13.1 & 49.1 & 63.7 & 21.8 \\
		SAN19-patch. & 78.2 & 12.1 & 55.1 & 62.0 & 24.8 \\
		\bottomrule[1pt]
	\end{tabular}
	\vspace{1mm}
	\caption{Robustness of trained networks to adversarial attacks on the val-original set. $n$ is the number of attack iterations. `s.\ rate' is the success rate of the attack (lower is better) and `top-1' is the accuracy under the attack (higher is better). Self-attention models are much more robust than convolutional networks.}
	\label{tab:attack}
	\vspace{-3mm}
\end{table}

Both experiments indicate that self-attention networks may have significant benefits in terms of robustness and generalization. These may surpass accuracy gains observed in traditional evaluation procedures and merit further study.

%% file: tex/conclusion.tex
In this paper, we explored the effectiveness of image recognition models that are based fully on self-attention. We considered two forms of self-attention: pairwise and patchwise. The pairwise form is a set operation and is fundamentally different from convolution in this respect. The patchwise form is a generalization of convolution. For both forms, we introduced \emph{vector} attention that efficiently adapts weights across both spatial dimensions and channels.

Our experiments yield a number of significant findings. First, networks based purely on pairwise self-attention match or outperform convolutional baselines. This indicates that the success of deep learning in computer vision is not inextricably tied to convolutional networks: there is an alternative route to comparable or higher discriminative power, with different and potentially beneficial structural properties such as permutation- and cardinality-invariance. Our second major finding is that patchwise self-attention models substantially outperform convolutional baselines. This suggests that patchwise self-attention, which generalizes convolution, may yield strong accuracy gains across applications in computer vision. Finally, our experiments indicate that \emph{vector} self-attention is particularly powerful and substantially outperforms scalar (dot-product) attention, which has been the predominant formulation to date.